\documentclass[11pt,letterpaper]{article}
\usepackage{emnlp2016}
\usepackage{times}
\usepackage{helvet}
\usepackage{courier}
\usepackage{multirow}
\usepackage{latexsym}
\usepackage{mathrsfs}
\usepackage{graphicx}
\usepackage{subfig}
\usepackage{amsfonts}
\usepackage{amssymb}
\usepackage{amsmath}
\usepackage{bm}
\usepackage{url}
\usepackage{tikz-dependency}
\usepackage{pgfplots}

\usepackage{xcolor}
\usepackage{booktabs}
\usetikzlibrary{positioning,bayesnet}
\allowdisplaybreaks

\newenvironment{itemize*}%
  {\begin{itemize}%
    \setlength{\itemsep}{0pt}%
    \setlength{\parskip}{1pt}}%
  {\end{itemize}}
  \newenvironment{enumerate*}%
  {\begin{enumerate}%
    \setlength{\itemsep}{0pt}%
    \setlength{\parskip}{1pt}}%
  {\end{enumerate}}

\pgfplotsset{compat=1.5}

\DeclareMathOperator*{\softmax}{softmax}

\def\W{\mathbf{W}}

\def\bb{\mathbf{b}}
\def\y{\mathbf{y}}

\def\h{\mathbf{h}}
\def\rr{\mathbf{r}}
\def\x{\mathbf{x}}
\def\cc{\mathbf{c}}
\def\ii{\mathbf{i}}
\def\ff{\mathbf{f}}
\def\oo{\mathbf{o}}

\def\g{\mathbf{g}}

\def\cc{\mathbf{c}}

\def\M{\mathbf{M}}
\def\r{\mathbf{r}}
\def\W{\mathbf{W}}

\def\bb{\mathbf{b}}

\def\h{\mathbf{h}}
\def\rr{\mathbf{r}}
\def\bx{\mathbf{x}}
\def\by{\mathbf{y}}
\def\cc{\mathbf{c}}
\def\ii{\mathbf{i}}
\def\ff{\mathbf{f}}
\def\oo{\mathbf{o}}

\def\g{\mathbf{g}}

\def\cc{\mathbf{c}}

\def\bx{\mathbf{x}}
\def\by{\mathbf{y}}
\emnlpfinalcopy 
\allowdisplaybreaks
\title{Deep Multi-Task Learning with Shared Memory}

\author{Pengfei Liu \quad Xipeng Qiu\thanks{\quad Corresponding author.} \quad Xuanjing Huang\\
Shanghai Key Laboratory of Intelligent Information Processing, Fudan University\\
School of Computer Science, Fudan University\\
825 Zhangheng Road, Shanghai, China\\
\{pfliu14,xpqiu,xjhuang\}@fudan.edu.cn}

\date{}

\begin{document}
\maketitle

\begin{abstract}
Neural network based models have achieved impressive results on various specific tasks. However, in previous works, most models are learned separately based on single-task supervised objectives, which often suffer from insufficient training data. In this paper, we propose two deep architectures which can be trained jointly on multiple related tasks. More specifically, we augment neural model with an external memory, which is shared by several tasks.
Experiments on two groups of text classification tasks show that our proposed architectures can improve the performance of a task with the help of other related tasks.
\end{abstract}

\section{Introduction}
Neural network based models have been shown to achieved impressive results on various NLP tasks rivaling or in some cases surpassing traditional models, such as text classification \cite{kalchbrenner2014convolutional,socher2013recursive,liu2015multitimescale}, semantic matching \cite{hu2014convolutional,liu2016deep}, parser \cite{chen2014fast} and machine translation \cite{bahdanau2014neural}.

Usually, due to the large number of parameters these neural models need a large-scale corpus.
It is hard to train a deep neural model that generalizes well with size-limited data, while
building the large scale resources for some NLP tasks is also a challenge.
To overcome this problem, these models often involve an unsupervised pre-training phase.  The final model is fine-tuned on specific task with respect to a supervised training criterion.
However, most pre-training methods are based on unsupervised objectives \cite{collobert2011natural,turian2010word,mikolov2013efficient}, which is effective to improve the final performance, but it does not directly optimize the desired task.

Multi-task learning is an approach to learn multiple related tasks simultaneously to significantly improve performance relative to learning each task independently.
Inspired by the success of multi-task learning \cite{caruana1997multitask},
several neural network based models \cite{collobert2008unified,liu2015representation} are proposed for NLP tasks, which   utilized multi-task learning to jointly learn several tasks with the aim of mutual benefit. The characteristic of these multi-task architectures is they share some lower layers to determine common features. After the shared layers, the remaining layers are split into multiple specific tasks.

In this paper, we propose two deep architectures of sharing information among several tasks in multi-task learning framework. All the related tasks are integrated into a single system which is trained jointly.
More specifically, inspired by Neural Turing Machine (NTM) \cite{graves2014neural} and memory network \cite{sukhbaatar2015end}, we equip task-specific long short-term memory (LSTM) neural network \cite{hochreiter1997long} with an external shared memory. The external memory has capability to store long term information and knowledge shared by several related tasks.
Different with NTM, we use a deep fusion strategy to integrate the information from the external memory into task-specific LSTM, in which a fusion gate controls the information flowing flexibly and enables the model to selectively utilize the shared information.

We demonstrate the effectiveness of our architectures on two groups of text classification tasks. Experimental results show that jointly learning of multiple related tasks can improve the performance of each task relative to learning them independently.

Our contributions are of three-folds:
\begin{itemize}
  \item We proposed a generic multi-task framework, in which different tasks can share information by an external memory and communicate by a reading/writing mechanism. Two proposed models are complementary to prior multi-task neural networks.
  \item Different with Neural Turing Machine and memory network, we introduce a deep fusion mechanism between internal and external memories, which helps the LSTM units keep them interacting closely without being conflated.
  \item As a by-product, the fusion gate enables us to better understand how the external shared memory helps specific task.
\end{itemize}

\section{Neural Memory Models for Specific Task}

In this section, we briefly describe LSTM model, and then propose an external memory enhanced LSTM with deep fusion.

\subsection{Long Short-term Memory}
Long short-term memory network (LSTM) \cite{hochreiter1997long} is a type of recurrent neural network (RNN) \cite{Elman:1990}, and
specifically addresses the issue of learning long-term dependencies.
LSTM maintains an internal memory cell that updates and exposes its content only when deemed necessary.

Architecturally speaking, the memory state and output state are explicitly separated by activation gates \cite{wang2015larger}. However, the limitation of LSTM is that it lacks a mechanism to index its memory while writing and reading \cite{ivo16}.

While there are numerous LSTM variants, here we use the LSTM architecture used by \cite{jozefowicz2015empirical}, which is similar to the architecture of \cite{graves2013generating} but without peep-hole connections.

We define the LSTM \emph{units} at each time step $t$ to be a collection of vectors in $\mathbb{R}^d$: an \emph{input gate} $\ii_t$, a \emph{forget gate} $\ff_t$,  an \emph{output gate} $\oo_t$, a \emph{memory cell} $\cc_t$ and a hidden state $\h_t$. $d$ is the number of the LSTM units. The elements of the gating vectors $\ii_t$, $\ff_t$ and $\oo_t$ are in $[0, 1]$.

The LSTM is precisely specified as follows.

\begin{align}
	\begin{bmatrix}
		\mathbf{\tilde{c}}_{t} \\
		\mathbf{o}_{t} \\
		\mathbf{i}_{t} \\
		\mathbf{f}_{t}
	\end{bmatrix}
	&=
	\begin{bmatrix}
		\tanh \\
		\sigma \\
		\sigma \\
		\sigma
	\end{bmatrix}
    \begin{pmatrix}
	\W_p
	\begin{bmatrix}
		\mathbf{x}_{t} \\
		\mathbf{h}_{t-1}
	\end{bmatrix}+\bb_p
    \end{pmatrix}, \label{eq:lstm1}\\
\mathbf{c}_{t} &=
		\mathbf{\tilde{c}}_{t} \odot \mathbf{i}_{t}
		+ \mathbf{c}_{t-1} \odot \mathbf{f}_{t}, \\
	\mathbf{h}_{t} &= \mathbf{o}_{t}  \odot \tanh\left( \mathbf{c}_{t}  \right)\label{eq:lstm3},
\end{align}
where $\bx_t \in \mathbb{R}^{m}$ is the input at the current time step;
$\W \in \mathbb{R}^{4h\times(d+m)}$ and $\bb_p \in \mathbb{R}^{4h}$ are parameters of affine transformation;
$\sigma$ denotes the logistic sigmoid function and $\odot$ denotes elementwise multiplication.

The update of each LSTM unit can be written precisely as follows:
\begin{align}
(\h_t,\cc_t) &= \mathbf{LSTM}(\h_{t-1},\cc_{t-1},\mathbf{x}_t, \theta_p).
\end{align}
Here, the function $\mathbf{LSTM}(\cdot, \cdot, \cdot, \cdot)$ is a shorthand for Eq. (\ref{eq:lstm1}-\ref{eq:lstm3}), and $\theta_p$ represents all the parameters of LSTM.

\subsection{Memory Enhanced LSTM}

LSTM has an internal memory to keep useful information for specific task, some of which may be beneficial to other tasks. However, it is non-trivial to share information stored in internal memory.

Recently, there are some works to augment LSTM with an external memory, such as neural Turing machine \cite{graves2014neural} and memory network \cite{sukhbaatar2015end}, called memory enhanced LSTM (ME-LSTM).  These models enhance the low-capacity internal memory to have a capability of modelling long pieces of text \cite{andrychowicz2016learning}.

Inspired by these models, we introduce an external memory to share information among several tasks. To better control shared information and understand how it is utilized from external memory, we propose a deep fusion strategy for ME-LSTM.

\begin{figure}[t]
  \centering
  \includegraphics[width=0.4\textwidth]{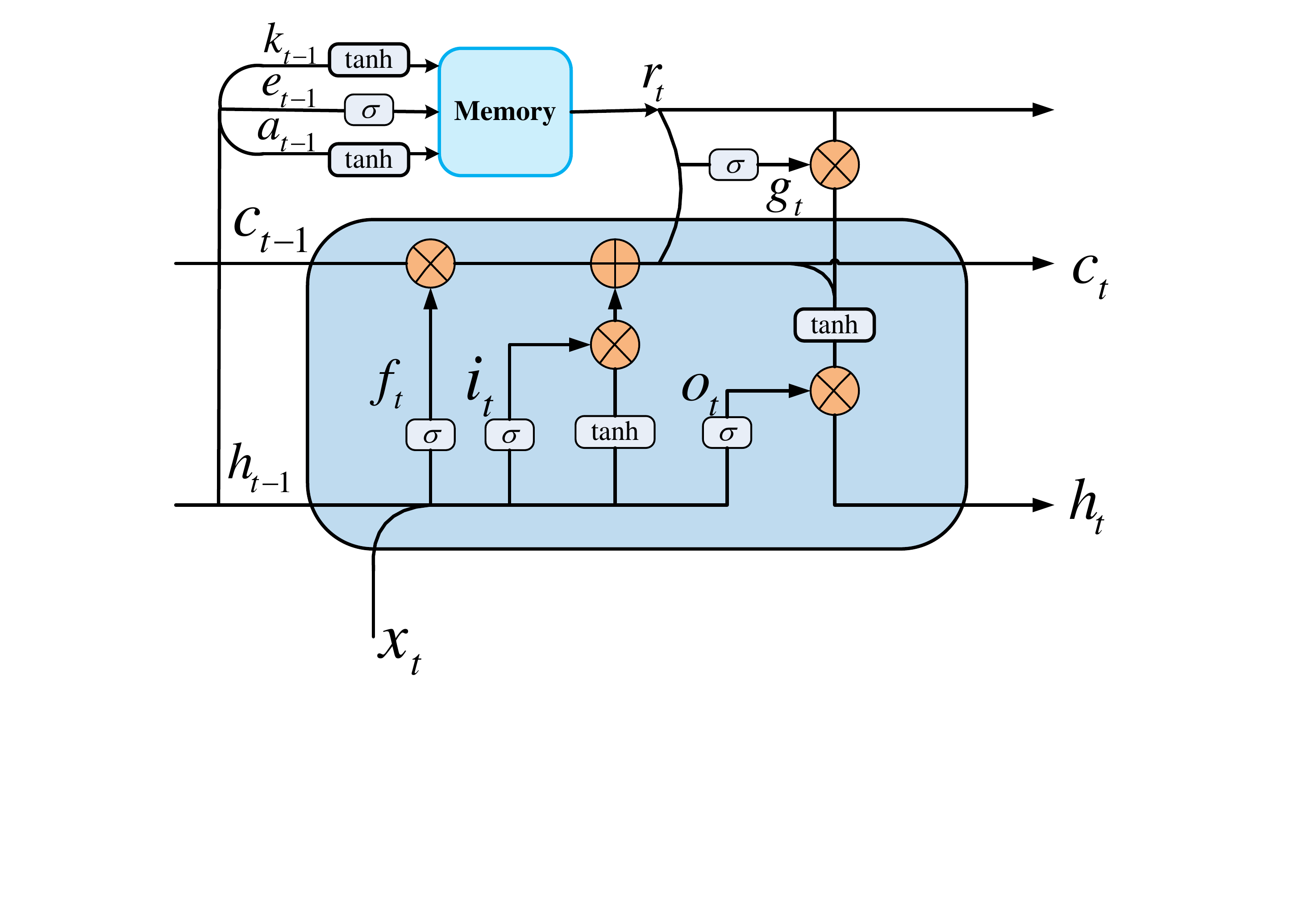}
  \caption{Graphical illustration of the proposed ME-LSTM unit with deep fusion of internal and external memories.
   }\label{fig:ME-LSTM}
\end{figure}

As shown in Figure \ref{fig:ME-LSTM}, ME-LSTM consists the original LSTM and an external memory which is maintained by reading and writing operations. The LSTM not only interacts with the input and output information but accesses the external memory using selective read and write operations.

The external memory and corresponding operations will be discussed in detail below.

\paragraph{External Memory}
The form of external memory is defined as a matrix $\M \in \mathbb{R}^{K\times M}$, where $K$ is the number of memory segments, and $M$ is the size of each segment. Besides, $K$ and $M$ are generally instance-independent and pre-defined as hyper-parameters.

At each step $t$, LSTM emits output $\h_t$ and three key vectors $\mathbf{k}_{t}$, $\mathbf{e}_{t}$ and $\mathbf{a}_{t}$ simultaneously. $\mathbf{k}_{t}$, $\mathbf{e}_{t}$ and $\mathbf{a}_{t}$  can be computed as
\begin{align}
	\begin{bmatrix}
		\mathbf{k}_{t} \\
		\mathbf{e}_{t} \\
		\mathbf{a}_{t} \\
	\end{bmatrix}
	=
	\begin{bmatrix}
		\tanh \\
		\sigma \\
		\tanh
	\end{bmatrix}
    (\W_m \mathbf{h}_{t} +\bb_m)
	\label{eq:ntm-emit}
\end{align}
where $\W_m$ and $\bb_m$ are parameters of  affine transformation.

\paragraph{Reading}
The read operation is to read information $\r_t \in \mathbb{R}^{M}$ from memory $\M_{t-1}$.
\begin{align}
 \r_t =  \alpha_{t} \M_{t-1},
\end{align}
where $\r_t$ denotes the reading vector and $\alpha_{t} \in \mathbb{R}^{K}$ represents a distribution over the set of segments of memory $\M_{t-1}$, which controls the amount of information to be read from and written to the memory.

Each scalar $\alpha_{t,k}$ in attention distribution $\alpha_t$ can be obtained as:
\begin{align}
\alpha_{t,k} &= \softmax(g(\M_{t-1,k},\mathbf{k}_{t-1}))
\end{align}
where $\M_{t-1,k}$ represents the $k$-th row memory vector, and $\mathbf{k}_{t-1}$ is a key vector emitted by LSTM.

Here $g(\mathbf{x},\mathbf{y})$ ($\mathbf{x} \in \mathbb{R}^{M},\mathbf{y} \in \mathbb{R}^{M}$) is a align function for which we consider two different alternatives:
\begin{equation}
    g(\bx,\by)=
   \begin{cases}
   \mathbf{v}^T \tanh(\W_a[\bx; \by]) \\
   \mathbf{cosine}(x,y)
   \end{cases}
\end{equation}
where $ \mathbf{v} \in \mathbb{R}^{M}$ is a parameter vector.

In our current implementation, the similarity measure is cosine similarity.

\paragraph{Writing}
The memory can be written by two operations: erase and add.
\begin{align}
        \M_t = \M_{t-1}(\mathbf{1}-\alpha_t\mathbf{e_t^T}) + \alpha_t\mathbf{a_t^T},
\end{align}
where $\mathbf{e}_{t},\mathbf{a}_{t} \in \mathbb{R}^{M}$ represent erase and add vectors respectively.

To facilitate the following statements, we re-write the writing equation as:
\begin{align}
    \M_t = \mathbf{f}_{write} ( \M_{t-1},\alpha_t, \h_t).
\end{align}

\paragraph{Deep Fusion between External and Internal Memories}

After we obtain the information from external memory, we need a strategy to comprehensively utilize information from both external and internal memory.

To better control signals flowing from external memory, inspired by \cite{wang2015larger}, we propose a deep fusion strategy to keep internal and external memories interacting closely without being conflated.

In detail, the state $\h_t$  of LSTM at step $t$ depends on both the read vector $\r_t$ from external memory, and internal memory $c_t$, which is computed by
\begin{align}
\h_t &=\oo_t \odot \tanh(\cc_t + \g_t \odot (\W_f \rr_t)),
\end{align}
where $\W_f$ is parameter matrix, and $\g_t$ is a fusion gate to select information from external memory, which is computed by
\begin{align}
\g_t = \sigma(\W_r \rr_t + \W_c \mathbf{c}_t),
\end{align}
where $\W_r$ and $\W_c$ are parameter matrices.

Finally, the update of external memory enhanced LSTM  unit can be written precisely as
\begin{align}
(\h_t,\M_t,\cc_t) = \mathbf{ME\textrm{-}LSTM}(\h_{t-1}, \nonumber\\
\M_{t-1}, \cc_{t-1}, \mathbf{x}_t, \theta_p, \theta_q),
\end{align}
where $\theta_p$ represents all the parameters of LSTM internal structure and $\theta_q$ represents all the parameters to maintain the external memory.

Overall, the external memory enables ME-LSTM to have larger capability to store more information, thereby increasing the ability of ME-LSTM. The read and write operations allow ME-LSTM to capture complex sentence patterns.

\begin{figure}[t]
  \centering \hspace{-3em}
  \subfloat[Global Memory Architecture]{
  \includegraphics[width=0.40\textwidth]{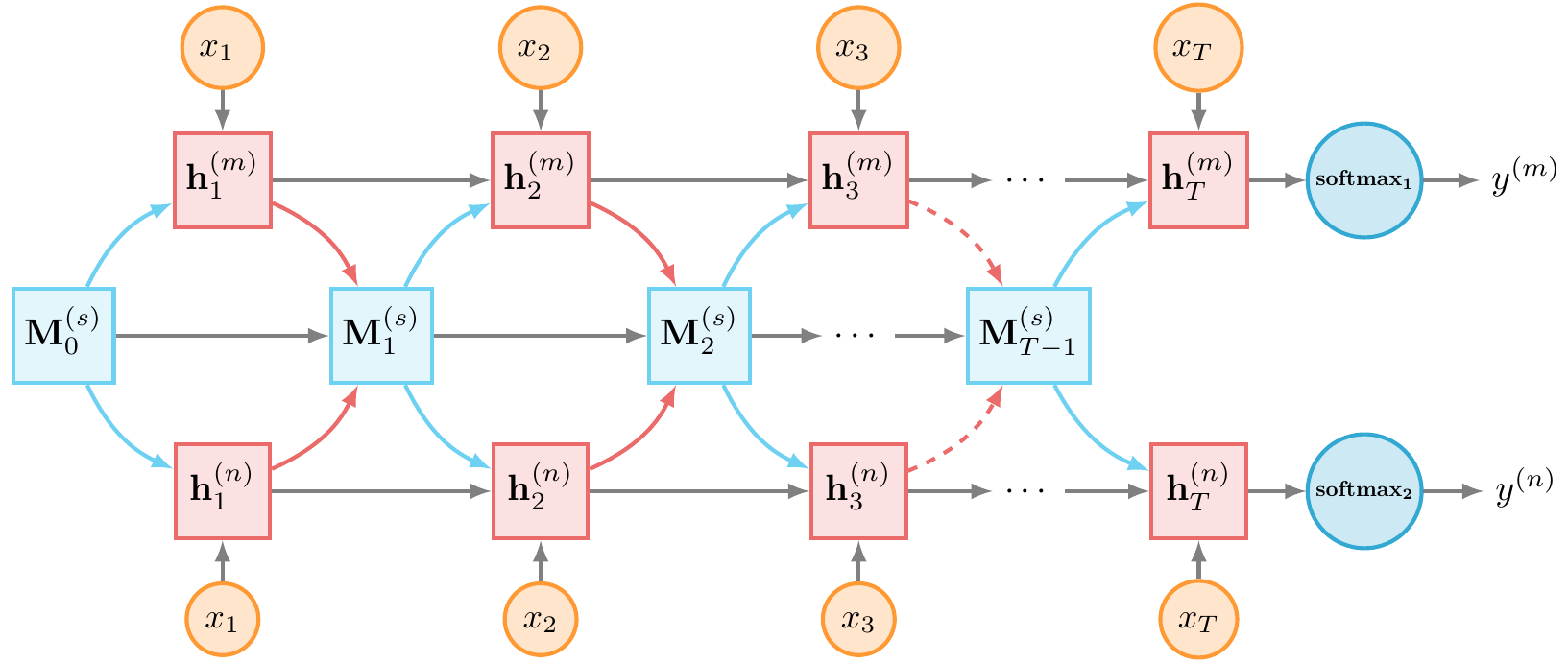}
  }\\\hspace{-3em}
  \subfloat[ Local-Global Hybrid Memory Architecture]{
  \includegraphics[width=0.42\textwidth]{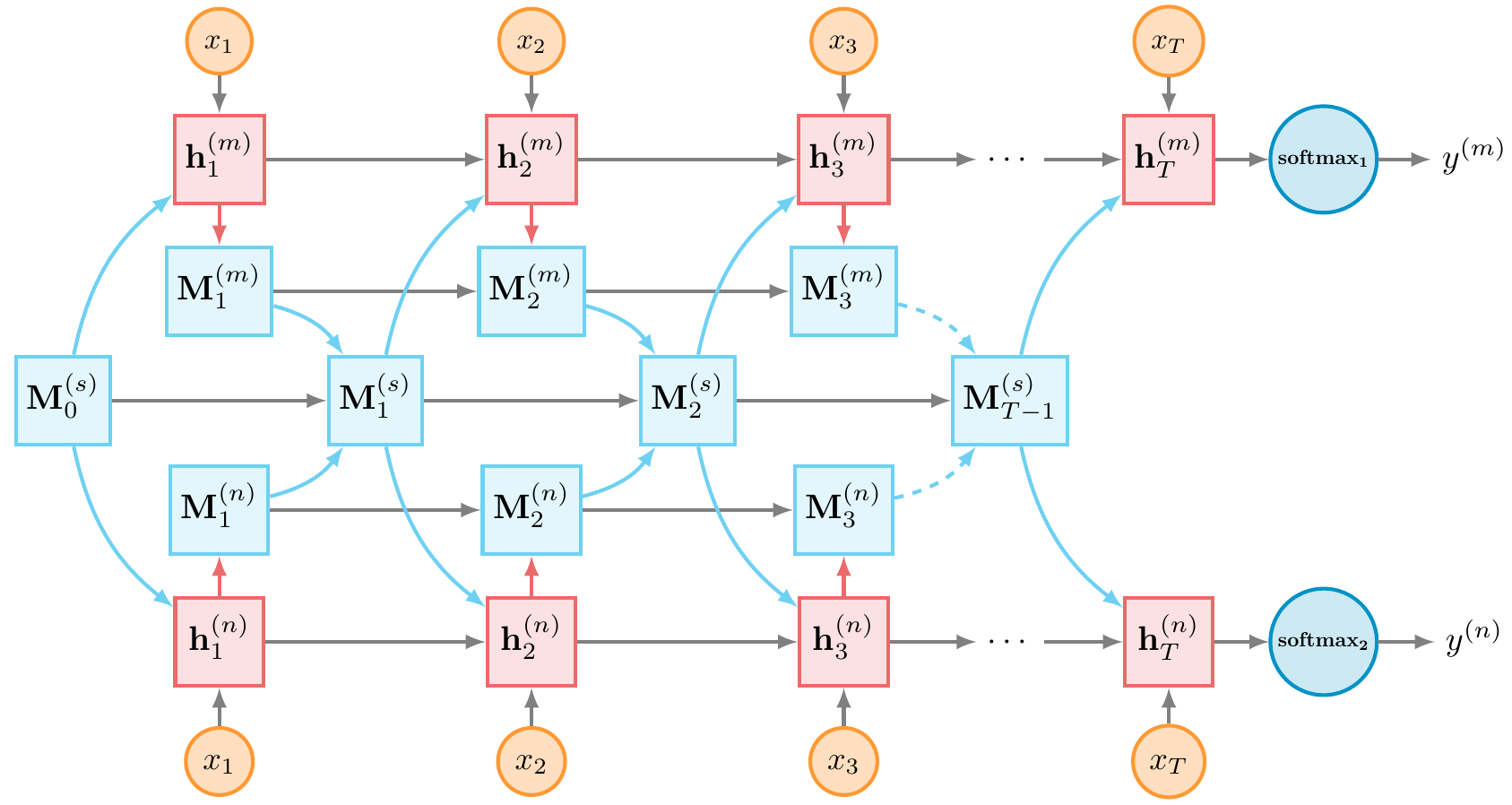}
  }
  \caption{Two architectures for modelling text with multi-task learning.
   }\label{fig:ours}
\end{figure}

\section{Deep Architectures with Shared Memory for Multi-task Learning}
Most existing neural network methods are based on supervised training objectives on a single task \cite{collobert2011natural,socher2013recursive,kalchbrenner2014convolutional}. These methods often suffer from the limited amounts of training data. To deal with this problem, these models often involve an unsupervised pre-training phase. This unsupervised pre-training is effective to improve the final performance, but it does not directly optimize the desired task.

Motivated by the success of multi-task learning \cite{caruana1997multitask}, we propose two deep architectures with shared external memory to leverage supervised data from many related tasks. Deep neural model is well suited for multi-task learning since the features learned from a task may be useful for other tasks. Figure \ref{fig:ours} gives an illustration of our proposed architectures.

\paragraph{ARC-I: Global Shared Memory}
In ARC-I, the input is modelled by a task-specific LSTM and external shared memory. More formally, given an input text $x$, the task-specific output $\h_{t}^{(m)}$ of task $m$ at step $t$ is defined as
\begin{align}
(\h_t^{(m)},\M_t^{(s)},\cc_t^{(m)}) = \mathbf{ME\textrm{-}LSTM}(\h_{t-1}^{(m)}, \nonumber\\
\M_{t-1}^{(s)}, \cc_{t-1}^{(m)}, \mathbf{x}_t, \theta_p^{(m)}, \theta_q^{(s)}),
\end{align}
where $\x_t$ represents word embeddings of word $x_t$; the superscript $s$ represents the parameters are shared across different tasks; the superscript $m$ represents that the parameters or variables are task-specific for task $m$.

Here all tasks share single global memory $\M^{(s)}$, meaning that all tasks can read information from
it and have the duty to write their shared or task-specific information into the memory.
\begin{align}
 \M_t^{(s)} &= \mathbf{f}_{write} ( \M_{t-1}^{(s)},\alpha_t^{(s)}, \h_t^{(m)} )。
\end{align}
After calculating the task-specific representation of text $\h_T^{(m)}$ for task $m$, we can predict the probability distribution over classes.

\begin{table*}[h!]\setlength{\tabcolsep}{3pt} \small
\centering
\begin{tabular}{|c|c|c|c|c|c|c|c|c|}
\hline

\multicolumn{2}{|c|}{\textbf{Dataset}} & \textbf{Type} & \textbf{Train Size} & \textbf{Dev. Size} & \textbf{Test Size} & \textbf{Class} & \textbf{Avg. Length} & \textbf{Vocabulary Size}  \\
\hline

\multirow{3}{*}{Movie} & SST-1 & Sen. & 8544 & 1101 & 2210 & 5 & 19 & 18K\\
& SST-2 & Sen. & 6920 & 872 & 1821 & 2 & 18 & 15K\\
& SUBJ & Sen. & 9000 & - & 1000 & 2 & 21 & 21K\\
& IMDB & Doc. & 25,000 & - & 25,000 & 2 & 294 & 392K\\
\hline

\multirow{4}{*}{Product} & Books & Doc. & 1400 & 200 & 400 & 2 & 181 & 27K \\
& DVDs & Doc. & 1400 & 200 & 400 & 2 & 197 & 29K \\
& Electronics & Doc. & 1400 & 200 & 400 & 2 & 117 & 14K \\
& Kitchen & Doc. & 1400 & 200 & 400 & 2 & 98 & 12K \\
\hline
\end{tabular}
\caption{Statistics of two multi-task datasets. Each dataset consists of four related tasks.} \label{tab:data}
\end{table*}

\paragraph{ARC-II: Local-Global Hybrid Memory}
In ARC-I, all tasks share a global memory, but can also record task-specific information besides shared information. To address this, we allocate each task a local task-specific external memory, which can further write shared information to a global memory for all tasks.

More generally, for task $m$, we assign each task-specific LSTM with a local memory $\M^{(m)}$, followed by a global memory $\M^{(s)}$, which is shared across different tasks.

The read and write operations of the local and global memory are defined as
\begin{align}
 \r_t^{(m)} &=  \alpha_{t}^{(m)} \M_{t}^{(m)}  , \\
 \M_t^{(m)} &= \mathbf{f}_{write} ( \M_{t-1}^{(m)},\alpha_t^{(m)}, \h_t^{(m)} ), \\
 \r_t^{(s)} &=  \alpha_{t-1}^{(s)} \M_{t-1}^{(s)} , \\
 \M_t^{(s)} &= \mathbf{f}_{write} ( \M_{t-1}^{(s)},\alpha_t^{(s)}, \r_t^{(m)} ),
\end{align}
where the superscript $s$ represents the parameters are shared across different tasks; the superscript $m$ represents that the parameters or variables are task-specific for task $m$.

In ARC-II, the local memories enhance the capacity of memorizing, while global memory enables the information flowing from different tasks to interact sufficiently.

\section{Training}

The task-specific representation $\h^{(m)}$, emitted by the deep muti-task architectures, is ultimately fed into the corresponding task-specific output layers.
\begin{align}
{\hat{\y}}^{(m)} = \softmax(\W^{(m)}\h^{(m)} + \bb^{(m)}),
\end{align}
where ${\hat{\y}}^{(m)}$ is prediction probabilities for task $m$.

Given $M$ related tasks, our global cost function is the linear combination of cost function for all tasks.
\begin{align}
\phi = \sum_{m=1}^{M}{\lambda}_m  L({\hat{y}}^{(m)}, y^{(m)})
\end{align}
where $\lambda_m$ is the weights for each task $m$ respectively.

\paragraph{Computational Cost}
Compared with vanilla LSTM, our proposed two models do not cause much extra computational cost while converge faster. In our experiment, the most complicated ARC-II, costs 2 times as long as vanilla LSTM.

\section{Experiment}
\begin{table}[t]  \setlength{\tabcolsep}{3pt} \small
\centering
\begin{tabular}{|l|*{2}{p{0.18\linewidth}|}}
    \hline
    & Movie Reviews & Product Reviews\\\hline
    Embedding dimension &100 & 100\\
    Hidden layer size &100 & 100\\
    External memory size & (50,20) & (50,20) \\
    Initial learning rate& 0.01 & 0.1\\
    Regularization & $0$ & $1E{-5}$\\
    \hline
\end{tabular}
\caption{Hyper-parameters of our models.}\label{tab:paramSet}
\end{table}

\begin{table*}[!t]\small
\center
\begin{tabular}{|l|*{7}{c|}}
\hline
\multicolumn{2}{|c|}{\textbf{Model}} &	 SST-1  & SST-2 &	SUBJ & IMDB &	 Avg$\Delta$\\
\hline
\multirow{2}{*}{Single Task}
 & LSTM     &   45.9    &   85.8    &   91.6    & 88.5    &  -   \\
 & ME-LSTM  &   46.4    &   85.5    &   91.0    & 88.7  &  -   \\
\hline
\multirow{4}{*}{Multi-task}

& ARC-I  &   48.6    &   87.0    &  93.8 & 89.8    &  +(1.8/1.9)    \\
& ARC-II &   \textbf{49.5}    &   \textbf{87.8}    & \textbf{95.0} &  \textbf{91.2}    & +(2.9/3.0)  \\
\cline{2-7}
& MT-CNN        &  46.7  & 86.1   & 92.2  &  88.4     &-\\
& MT-DNN     &	 44.5  & 84.0   & 90.1   &  85.6     &-\\

\hline
\multicolumn{2}{|c|}{NBOW} & 42.4 & 80.5 & 91.3 & 83.6 &-\\
\multicolumn{2}{|c|}{RAE \cite{socher2011semi}}   &	 43.2 &	 82.4 &	- & -& - \\
\multicolumn{2}{|c|}{MV-RNN \cite{socher2012semantic}} &	 44.4 &	 82.9 & - & -& -\\
\multicolumn{2}{|c|}{RNTN \cite{socher2013recursive}}  &	 45.7 &	 85.4 &	 - &- &-\\
\multicolumn{2}{|c|}{DCNN \cite{kalchbrenner2014convolutional}}  &	 48.5&	 86.8 &	 - &89.3 &-\\
\multicolumn{2}{|c|}{CNN-multichannel \cite{kim2014convolutional}}  &	47.4 & \textbf{88.1}& 93.2 &- &-\\
\multicolumn{2}{|c|}{Tree-LSTM \cite{tai2015improved}}  &	 \textbf{50.6}&	 86.9 & - &- &-\\
\hline
\end{tabular}
\caption{
Accuracies of our models on  movie reviews tasks against state-of-the-art neural models. The last column gives the  improvements relative to LSTM and ME-LSTM respectively.
\textbf{NBOW}: Sums up the word vectors and applies a non-linearity followed by a softmax classification layer.  \textbf{RAE}: Recursive Autoencoders with pre-trained word vectors from Wikipedia \protect\cite{socher2011semi}.
\textbf{MV-RNN}: Matrix-Vector Recursive Neural Network with parse trees \protect\cite{socher2012semantic}.
\textbf{RNTN}: Recursive Neural Tensor Network with tensor-based feature function and parse trees \protect\cite{socher2013recursive}.
\textbf{DCNN}: Dynamic Convolutional Neural Network with dynamic k-max pooling \protect\cite{kalchbrenner2014convolutional,denil2014modelling}.
\textbf{CNN-multichannel}: Convolutional Neural Network \protect\cite{kim2014convolutional}.
\textbf{Tree-LSTM}: A generalization of LSTMs to tree-structured network topologies \protect\cite{tai2015improved}.
}\label{tab:result-movie}
\end{table*}

In this section, we investigate the empirical performances of our proposed architectures on two multi-task datasets. Each dataset contains several related tasks.

\subsection{Datasets}

The used multi-task datasets are briefly described as follows.  The detailed statistics are listed in Table \ref{tab:data}.

\paragraph{Movie Reviews}
The movie reviews dataset consists of four sub-datasets about movie reviews.
\begin{itemize*}
  \item \textbf{SST-1} The movie reviews with five classes in the Stanford Sentiment Treebank\footnote{\url{http://nlp.stanford.edu/sentiment}.} \cite{socher2013recursive}.
  \item \textbf{SST-2} The movie reviews with binary classes. It is also from the Stanford Sentiment Treebank.
  \item \textbf{SUBJ} The movie reviews with labels of subjective or objective \cite{pang2004sentimental}.
  \item \textbf{IMDB} The IMDB dataset\footnote{\url{http://ai.stanford.edu/~amaas/data/sentiment/}} consists of 100,000 movie reviews with binary classes  \cite{maas2011learning}. One key aspect of this dataset is that each movie review has several sentences.
\end{itemize*}

\paragraph{Product Reviews}
This dataset\footnote{\url{https://www.cs.jhu.edu/~mdredze/datasets/sentiment/}}, constructed by \newcite{blitzer2007biographies}, contains Amazon product reviews from four different domains: Books, DVDs, Electronics and Kitchen appliances. The goal in each domain is to classify a product review as either positive or negative.
The datasets in each domain are partitioned randomly into
training data, development data and testing data with the proportion of 70\%, 20\% and 10\% respectively.

\subsection{Competitor Methods for Multi-task Learning}
The multi-task frameworks proposed by previous works are various while not all can be applied to the tasks we focused. Nevertheless, we chose two most related neural models for multi-task learning and implement them as strong competitor methods .
\begin{itemize}
    \item MT-CNN: This model is proposed by \newcite{collobert2008unified} with convolutional layer, in which lookup-tables are shared partially while other layers are task-specific.
    \item MT-DNN: The model is proposed by \newcite{liu2015representation} with bag-of-words input and multi-layer perceptrons, in which a hidden layer is shared.
\end{itemize}

\subsection{Hyperparameters and Training}
The networks are trained with backpropagation and the gradient-based optimization is performed using the Adagrad update rule \cite{duchi2011adaptive}.

The word embeddings for all of the models are initialized with the 100d GloVe vectors (840B token version, \cite{pennington2014glove}) and fine-tuned during training to improve the performance.
The other parameters are initialized by randomly sampling from uniform distribution in $[-0.1, 0.1]$.
The mini-batch size is set to 16.

For each task, we take the hyperparameters which achieve the best performance on the development set via an small grid search over combinations of the initial learning rate $[0.1, 0.01]$, $l_2$ regularization $[0.0, 5E{-5}, 1E{-5}]$. For datasets without development set, we use 10-fold cross-validation (CV) instead.
The final hyper-parameters are set as Table \ref{tab:paramSet}.

\subsection{Multi-task Learning of  Movie Reviews}
We first compare our proposed models with the baseline system for single task classification.
Table \ref{tab:result-movie} shows the classification accuracies on the movie reviews dataset. The row of ``Single Task'' shows the results of LSTM and ME-LSTM for each individual task. With the help of multi-task learning, the performances of these four tasks are improved by 1.8\% (ARC-I) and  2.9\% (ARC-II) on average relative to LSTM. We can find that the architecture of local-global hybrid external memory has better performances. The reason is that the global memory in ARC-I could store some task-specific information besides shared information, which maybe noisy to other tasks.
Moreover, both of our proposed models outperform MT-CNN and MT-DNN, which indicates the effectiveness of our proposed shared mechanism.
To give an intuitive evaluation of these results, we also list the following state-of-the-art neural models. With the help of utilizing the shared information of several related tasks, our results outperform most of state-of-the-art models.
Although Tree-LSTM outperforms our method on SST-1, it needs an external parser to get the sentence topological structure. It is worth noticing that our models are generic and compatible with the other LSTM based models. For example, we can easily extend our models to incorporate the Tree-LSTM model.

\begin{table*}[!t]
\center
\begin{tabular}{|l|*{7}{c|}}
\hline
\multicolumn{2}{|c|}{\textbf{Model}} &	 Books  & DVDs &	Electronics &	 Kitchen  & Avg$\Delta$\\
\hline
\multirow{2}{*}{Single Task}
& LSTM     &  78.0    &   79.5    &   81.2   &   81.8    & - \\
& ME-LSTM  &   77.5    &   80.2    &   81.5    & 82.1  & -   \\
\hline
\multirow{4}{*}{Multi-task}
& ARC-I  &   {81.2}    &   82.0    &   84.5    &   \textbf{84.3} & +(2.9/2.6)   \\
& ARC-II &   \textbf{82.8}    &   \textbf{83.0}    &   \textbf{85.5}    &   {84.0} & +(3.7/3.5)   \\
\cline{2-7}
& MT-CNN        & 80.2   & 81.0          & 83.4    & 83.0 &-\\
& MT-DNN       &79.7    &80.5           & 82.5    &82.8 &-\\

\hline
\end{tabular}
\caption{Accuracies of our models on product reviews dataset. The last column gives the  improvement relative to LSTM and ME-LSTM respectively.}\label{tab:result-product}
\end{table*}

\subsection{Multi-task Learning of Product Reviews}

Table \ref{tab:result-product} shows the classification accuracies on the tasks of product reviews. The row of ``Single Task'' shows the results of the baseline for each individual task. With the help of global shared memory (ARC-I), the performances of these four tasks are improved by an average of 2.9\%(2.6\%) compared with LSTM(ME-LSTM).  ARC-II achieves best performances on three sub-tasks, and its average improvement is 3.7\%(3.5\%).
Compared with MT-CNN and MT-DNN, our models achieve a better performance. We think the reason is that our models can not only share lexical information but share complicated patterns of sentences by reading/writing operations of external memory.
Furthermore, these results on product reviews are consistent with that on movie reviews, which shows our architectures are robust.

\begin{figure*}[!t]
\centering
   \subfloat[]{
   \includegraphics[width=0.38\textwidth,height=6em]{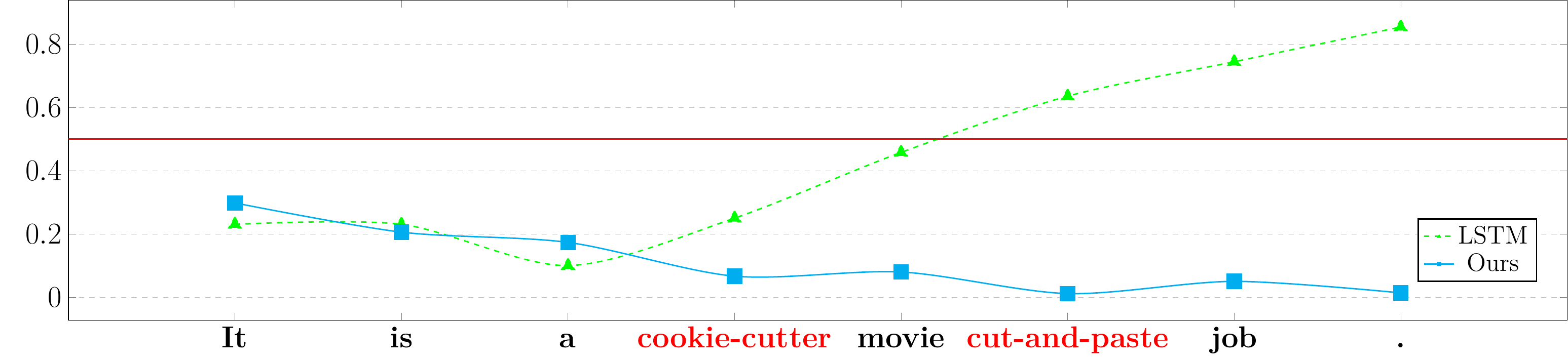}
   }
   \subfloat[]{
   \includegraphics[width=0.59\textwidth,height=6em]{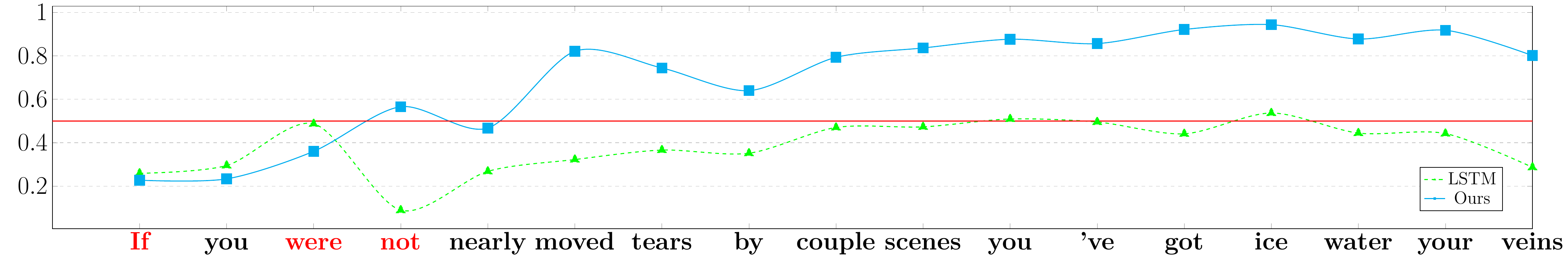}
   }

    \subfloat[]{
   \includegraphics[width=0.38\textwidth,height=6em]{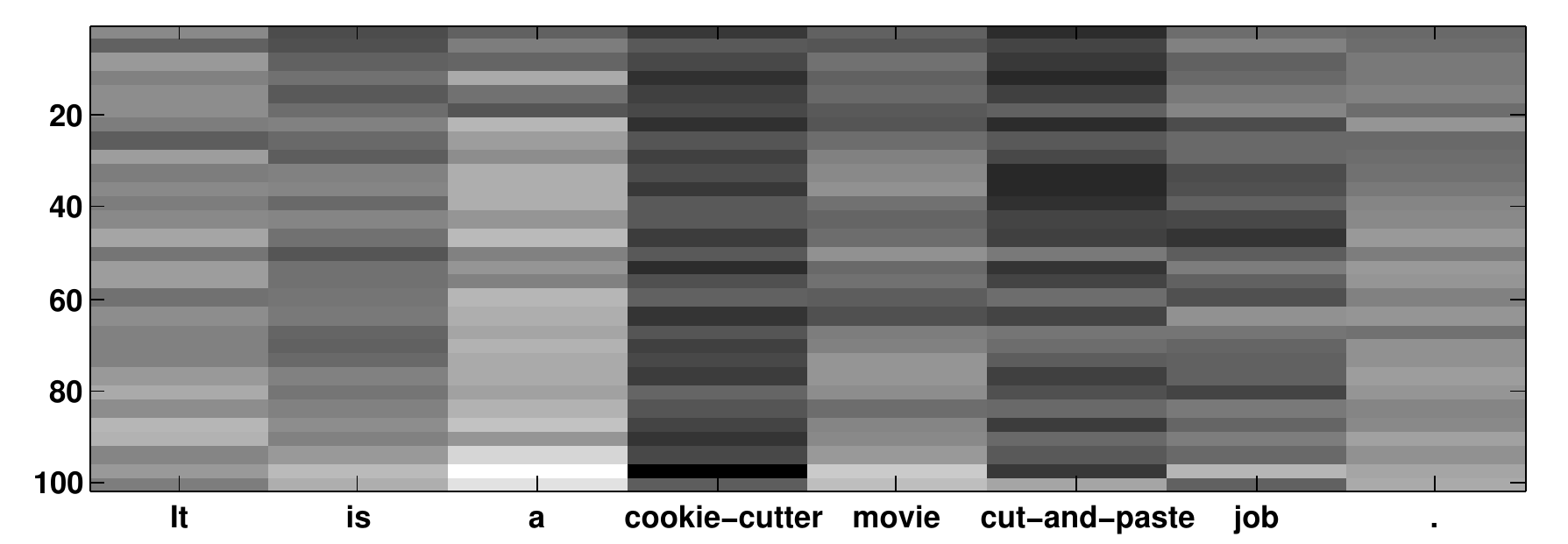}
   }
    \subfloat[]{
   \includegraphics[width=0.58\textwidth,height=6em]{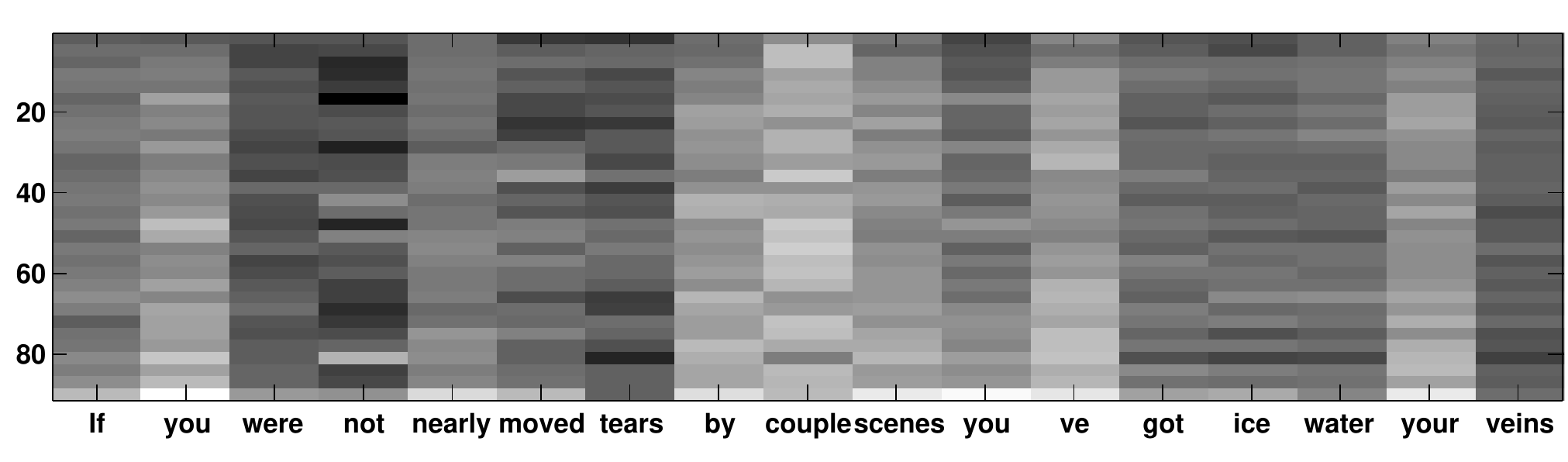}
   }

  \caption{(a)(b) The change of the predicted sentiment score at different time steps. Y-axis represents the sentiment score, while X-axis represents the input words in chronological order. The red horizontal line gives a border between the positive and negative sentiments. (c)(d) Visualization of the fusion gate's  activation.
}\label{fig:exp-case}
\end{figure*}

\subsection{Case Study}

To get an intuitive understanding of what is happening when we use shared memory to predict the class of text, we design an experiment to compare and analyze the difference between our models and vanilla LSTM, thereby demonstrating the effectiveness of our proposed architectures.

We sample two sentences from the SST-2 validation dataset, and the changes of the predicted sentiment score at different time steps are shown in Figure \ref{fig:exp-case}, which are obtained by vanilla LSTM and ARC-I respectively. Additionally, both models are bidirectional for better visualization. To get more insights into how the shared external memory influences the specific task, we plot and observe the evolving activation of fusion gates through time, which controls signals flowing from a shared external memory  to task-specific output, to understand the behaviour of neurons.

For the sentence ``\textit{It is a cookie-cutter movie, a cut-and-paste job.}'', which has a negative sentiment, while the standard LSTM gives a wrong prediction due to not understanding the informative words ``\textit{cookie-cutter}'' and ``\textit{cut-and-paste}''.

In contrast, our model makes a correct prediction and the reason can be inferred from the activation of fusion gates.  As shown in Figure \ref{fig:exp-case}-(c), we can see clearly the neurons are activated much when they take input as ``\textit{cookie-cutter}'' and ``\textit{cut-and-paste}'', which indicates  much information in shared memory has be passed into LSTM, therefore enabling the model to give a correct prediction.

Another case ``\textit{If you were not nearly moved to tears by a couple of scenes , you 've got ice water in your veins}'', a subjunctive clause introduced by ``\textit{if}'', has a positive sentiment.

As shown in Figure \ref{fig:exp-case}-(b,d), vanilla LSTM failed to capture the implicit meaning behind the sentence, while our model is sensitive to the pattern ``\textit{If ... were not ...}'' and has an accurate understanding of the sentence, which indicates the shared memory mechanism can not only enrich the meaning of certain words, but teach some information of sentence structure to specific task.

\section{Related Work}

Neural networks based multi-task learning has been proven effective in many NLP problems \cite{collobert2008unified,glorot2011domain,liu2015representation,liu2016recurrent}. In most of these models, the lower layers are shared across all tasks, while top layers are task-specific.

\newcite{collobert2008unified} used a shared representation for input words and solved different traditional NLP tasks within one framework. However, only one lookup table is shared, and the other lookup tables and layers are task-specific.

\newcite{liu2015representation} developed a multi-task DNN for learning representations across multiple tasks. Their multi-task DNN approach combines tasks of query classification and ranking for web search. But the input of the model is bag-of-word representation, which loses the information of word order.

More recently, several multi-task encoder-decoder networks were also proposed for  neural machine translation \cite{dong2015multi,luong2015multi,firat2016multi}, which can make use of cross-lingual information.

Unlike these works, in this paper we design two neural architectures with shared memory for multi-task learning, which can store useful information across the tasks. Our architectures are relatively loosely coupled, and therefore more flexible to expand. With the help of shared memory, we can obtain better task-specific sentence representation by utilizing the knowledge obtained by other related tasks.

\section{Conclusion and Future Work}

In this paper, we introduce two deep architectures for multi-task learning. The difference with the previous models is the mechanisms of sharing information among several tasks. We design an external memory to store the knowledge shared by several related tasks.
Experimental results show that our models can improve the performances of several related tasks by exploring common features.

In addition, we also propose a deep fusion strategy to integrate the information from the external memory into task-specific LSTM with a fusion gate.

In future work, we would like to investigate the other sharing mechanisms of neural network based multi-task learning.

\section*{Acknowledgments}
We would like to thank the anonymous reviewers for their valuable comments. This work was partially funded by National Natural Science Foundation of China (No. 61532011 and 61672162), the National High Technology Research and Development Program of China (No. 2015AA015408).

\bibliographystyle{emnlp2016}
\bibliography{nlp2,ours,nlp}
\end{document}